\documentclass[conference]{IEEEtran}
\usepackage{enumerate}
\usepackage{enumitem}
\usepackage{tabulary}
\usepackage{xcolor}
\usepackage{comment}
\usepackage{enumerate}
\usepackage{enumitem}
\usepackage{graphicx}
\usepackage{dsfont}
\usepackage{amssymb}
\usepackage{mathtools}
\usepackage{wrapfig}
\usepackage{multirow}
\usepackage{hhline}
\usepackage{amsmath}
\usepackage{hyperref}

\newcommand{\treeDA}{${\mbox{\tiny Tr12A}}$}

\newcommand{\gbmmDA}{${\mbox{\tiny gbmm12A}}$}

\newcommand{\de}{${\mbox{\texttt{DE}}}$}
\newcommand{\pde}{${\cal P}^{\mbox{\texttt{DE}}}$}

\newcommand{\detreeDA}{${\cal P}^{\mbox{\texttt{DE}}}_{\mbox{\tiny Tr12A}}$}

\newcommand{\degbmmDA}{${\cal P}^{\mbox{\texttt{DE}}}_{\mbox{\tiny gbmm12A}}$}

\newcommand{\ga}{${\mbox{\texttt{GA}}}$}
\newcommand{\pga}{${\cal P}^{\mbox{\texttt{GA}}}$}

\newcommand{\gatreeDA}{${\cal P}^{\mbox{\texttt{GA}}}_{\mbox{\tiny Tr12A}}$}

\newcommand{\gagbmmDA}{${\cal P}^{\mbox{\texttt{GA}}}_{\mbox{\tiny gbmm12A}}$}

\newcommand{\pso}{${\mbox{\texttt{PSO}}}$}
\newcommand{\ppso}{${\cal P}^{\mbox{\texttt{PSO}}}$}

\newcommand{\psotreeDA}{${\cal P}^{\mbox{\texttt{PSO}}}_{\mbox{\tiny Tr12A}}$}

\newcommand{\psogbmmDA}{${\cal P}^{\mbox{\texttt{PSO}}}_{\mbox{\tiny gbmm12A}}$}

\newcommand{\gad}{${\mbox{\texttt{GAD}}}$}
\newcommand{\pgad}{${\cal P}^{\mbox{\texttt{GAD}}}$}

\newcommand{\gadtreeDA}{${\cal P}^{\mbox{\texttt{GAD}}}_{\mbox{\tiny Tr12A}}$}

\newcommand{\gadgbmmDA}{${\cal P}^{\mbox{\texttt{GAD}}}_{\mbox{\tiny gbmm12A}}$}

\newcommand{\htreeDA}{${\cal P}^{\mbox{\tiny Het}}_{\mbox{\tiny Tr12A}}$}
\newcommand{\rhtreeDA}{${\cal P}^{\mbox{\tiny recHet}}_{\mbox{\tiny Tr12A}}$}
\newcommand{\hgbmmDS}{${\cal P}^{\mbox{\tiny Het}}_{\mbox{\tiny gbmm12S}}$}
\newcommand{\hgbmmDA}{${\cal P}^{\mbox{\tiny Het}}_{\mbox{\tiny gbmm12A}}$}
\newcommand{\rhgbmmDA}{${\cal P}^{\mbox{\tiny recHet}}_{\mbox{\tiny gbmm12A}}$}

\title{Reconfigurable Heterogeneous Parallel Island Models}

\author{\IEEEauthorblockN{Lucas A. da Silveira$^\dag$, Thaynara A. de Lima$^\ddag$ and Mauricio Ayala-Rinc\'on$^\dag$}
           \IEEEauthorblockA{
     $^\dag$Department of Computer Science, Universidade de Bras\'ilia,  
     70900--010 Bras\'ilia D.F., Brazil \\
     $^\ddag$Institute of Mathematics and Statistics, Universidade
     Federal de Goi\'as | Campus II, 
     74690--900 Goi\^ania, Brazil } }

\date{}

\begin{document}
\maketitle
\begin{abstract}
Heterogeneous Parallel Island Models (HePIMs) run different bio-inspired algorithms (BAs) in their islands. From a variety of communication topologies and migration policies fine-tuned for homogeneous PIMs (HoPIMs), which run the same BA in all their islands, previous work introduced HePIMs that provided competitive quality solutions regarding the best-adapted BA in HoPIMs. This work goes a step forward, maintaining the population diversity provided by HePIMs, and increasing their flexibility, allowing BA \textit{reconfiguration} on islands during execution: according to their performance, islands may substitute their BAs dynamically during the evolutionary process. Experiments with the introduced architectures (RecHePIMs) were applied to the NP-hard problem of sorting permutations by reversals,  using four different BAs, namely, simple Genetic Algorithm (\ga), Double-point crossover Genetic Algorithm (\gad), Differential Evolution (\de), and self-adjusting Particle Swarm Optimization (\pso). The results showed that the new reconfigurable heterogeneous models compute better quality solutions than the HePIMs closing the gap with the HoPIM running the best-adapted BA.
\end{abstract}

\section{Introduction}

Evolutionary computing relies on the evolution of candidate solutions over a finite number of generations to obtain accurate solutions for complex optimization problems. Genetic Algorithms (GAs) and Particle Swarm Optimization (PSO), among other  Bio-inspired algorithms (BAs), have been applied to successfully solve a diversity of such problems applied in engineering and sciences.   BAs guide the evolution of a population of individuals (candidate solutions) to improve their fitness to achieve a feasible solution to the target problem. BAs apply specialized computational rules to promote individual information exchange to the benefit of the population. However, optimization through such BA approaches demands exhaustive computations and efficient resources. Parallelization is one of the natural mechanisms applied to speed up and improve the accuracy of solutions obtained by BAs. 

This work studies Parallel Island Models (PIMs) that partition the population among their islands (processors)  and simultaneously run a BA in each island. In addition to the exchange of individuals, PIMs promote migration between islands. When all islands run the same BA, such models are called homogeneous PIMs (HoPIMs). This work improves heterogeneous PIMs (HePIMs) \cite{Lucas2021}, in which islands may run different BAs, allowing algorithmic \textit{reconfiguration} on their islands, i.e., islands may dynamically update their BAs. In addition to an adequate and well-calibrated migration policy required by HePIMs, reconfigurable HePIMs exchange information between islands to decide how islands should reconfigure their BAs.   

Silveira {\em et al.} \cite{Lucas2021} introduced reconfigurable HePIMs running four different BAs in their islands, namely, Genetic Algorithm (\ga), double-point crossover GA (\gad), Differential Evolution (\de), and self-adjusting Particle Swarm Optimization (\pso)  (see e.g., \cite{Holland1973}, \cite{DE1970}, and \cite{eberhart1995particle}).  PIMs performance depends on a good calibration of the breeding cycle parameters (related to the involved BA) and, in addition to vital aspects of the parallel island architecture as island communication synchronism, island migration frequency, communication topology, and migration policy.  We select two successful asynchronous HePIMs from \cite{Lucas2021} maintaining their parameters and adding the reconfiguration frequency.

The new reconfigurable HePIMs are tested in solving the unsigned reversal distance problem (URD), an ${\cal NP}$-hard problem (\cite{Caprara1997sorting}). Approaches to solve URD are applied in combinatorics to explain algebraic aspects of permutations (\cite{DELIMA201859}) and, in genomics, to analyze the evolutionary distance between genomes (\cite{kececioglu1993exact}), also represented as permutations.   

\vspace{2mm}
\noindent{\bf Main contributions.}
We design new reconfigurable HePIMs that run \ga, \gad, \de, and \pso\ in their islands using two successful asynchronous topologies from \cite{Lucas2021}. Non-reconfigurable HePIMs computed competitive solutions regarding most HoPIMs. The new reconfigurable architectures showed promising results computing solutions that exceed the quality of pure HePIMs and are very competitive regarding the best adapted HoPIMs, namely, the ones that run \de\ in all their islands. The heterogeneity of the new model effectively shares, through the migration policy, the good results experimented by individuals guided by different BAs in each island to the whole architecture. Furthermore, the reconfiguration ability shares the good experiences of the BAs in each island of the model. Adding the reconfiguration capability, the new model exceeds the flexibility of HoPIMs (all islands may update their BA to a unique BA) and of HePIMs (reconfiguration may update island BAs to the fix configuration of any non-reconfigurable HePIM) 
 
\vspace{2mm} 

\noindent{\bf Organization.} 
Sec. \ref{sec:background} discusses PIMs, the unsigned reversal distance problem, and the four selected BAs: \ga, \gad, \de, and \pso.  Sec. \ref{sec:topologies} introduces the new reconfigurable HePIMs explaining how different BAs are reconfigured.  Then,   Sec. \ref{sec:experimentsaccuracy} presents experiments and discusses accuracy results and statistical analysis.  Finally, Sec. \ref{sec:relatedwork} present related work before Sec.  \ref{sec:conclusion} that concludes and discussed  future work.  Source and data used in the experiments are available at \href{http://genoma.cic.unb.br}{http://genoma.cic.unb.br}.

\section{Background}\label{sec:background}

\subsection{Parallel island model (PIM)}\label{ssec:pim}

PIMs were proposed initially for GAs \cite{Crainic2003} and, besides improving speed-up, it is expected that such models also boost the solutions provided by sequential \ga.

The population is distributed into islands whose number is determined by the developer, running their BAs in parallel. 
The connection between the islands establishes the model's topology. \textit{Static} PIMs maintain the connections fixed during the execution, whereas \textit{dynamic} models admit changes during the process. 
Linked islands exchange individuals to evolve. Such a transfer can be uni- or bi-directionally. Different topologies and strategies to implement them are available (e.g. \cite{Duarte2020,Lucas2020,Sudholt2015parallel}).
\textit{Homogeneous} PIMs execute the same BA in all islands, whereas \textit{heterogeneous} models admit different BAs running in their islands. 
Figure \ref{fig:heterogeneousBTree} illustrates an heterogeneous static bi-directional tree topology. The edges show the connections between islands and remain unchanged, while vertices represent the islands with their BAs.

A \textit{migration policy} guides the exchange of individuals between islands during the evolutionary process.
PIMs have breeding-cycle and migration parameters tuned to improve the quality of solutions. In the following, the migration parameters are briefly presented. Some of them consider the classification of individuals as {\bf best}, {\bf worst} and {\bf random}, based on a rank established according to their fitness. The first half corresponds to the best and the second to the worst individuals, whereas random individuals are selected randomly.

\begin{itemize}%[leftmargin=1mm,itemsep=0em]

\item  Individuals number ({\it IN}):  number of individuals emigrating from each island.

\item Emigrant Individuals ({\it EMI}): rule the type of individuals selected for emigration among:
1. {\bf best}, 2. {\bf worst}, and 3. {\bf random}.
		   
\item Immigrant Individuals ({\it IMI}): determines the type of individuals in the target island replaced by immigrants among:
1. {\bf worst}, 2. {\bf random}, and 3. {\bf similar}. Similar individuals have the same classification as their replacement immigrants according to the fitness rank. 

\item Emigration Policy  ({\it EP}): defines whether individuals are {\bf cloned} or {\bf removed} in the local island when they emigrate to the target island.
\item Migration Interval ({\it MI}): corresponds to a percentage of iterations of the evolutionary process, called generations, after which the migration process is redone. 
Each island separately evolves its population by $\textit{MI} \times \textit{maxIt}$ generations, where \textit{maxIt} is the total number of iterations performed by each BA.
		  
		\end{itemize}

 PIMs are classified according to the synchroneity in which islands evolve their population. In \textit{Synchronous} PIMs, islands evolve performing each generation simultaneously, whereas, in \textit{asynchronous} PIMs, islands evolve independently even during migration. The latest mimics the behavior found in nature.

Here, we introduce reconfigurable heterogeneous PIMs. First, at a fixed generation percentage, called {\it Reconfiguration Frequency (RF)}, it is checked what islands have the best and the worst solutions regarding a metric based on the fitness average and the variance of the population. Then, the worst island updates its BA to the BA applied by the best island.

\subsection{Case-study}\label{subsec:case}

The evolutionary distance between two organisms can be computed as the number of rearrangements needed to transform a genome into another one by using some evolutionary measure. 
The authors consider the minimum number of reversals to compute the distance between unichromosomal organisms in this work. 

Permutations on $\{1,\cdots,n \}$ represent a genome containing $n$ genes.
Given a genome $\pi=(\pi_1, \pi_2, ..., \pi_n)$, where $1\leq i, \pi_i\leq n$, a reversal $\rho^{j,k}$, for  $1\leq j \leq k \leq n$, transforms $\pi$ into $\pi'=(\cdots, \pi_{j-1},\pi_k,\cdots,\pi_j, \pi_{k+1},\cdots)$, that is, it inverts the elements between $\pi_j$ and $\pi_k$.
If the orientation of the genes is known, each one receives a positive or negative sign, and the genome is a signed permutation. There are two evolutionary problems related to computing the distance by reversals. The signed reversal distance (SRD) problem asks for the minimum number of reversals needed to transform a signed permutation into another. On the other hand, the unsigned reversal distance (URD) problem consists of computing such a number between unsigned permutations, which orientation of genes is unknown. It is well-known that SRD belongs to class $\cal P$ \cite{Hannenhall1999}, whereas URD is an ${\cal NP}$-hard problem \cite{Caprara1997sorting}.

Our models are applied to solve URD. The fitness used by the algorithms is computed over signed permutations, generated after a random assignment of signs to each gene of a permutation.

\subsection{Local Evolutionary Engines | bio-inspired Algorithms}
	 
Four BAs, widely used for analyzing optimization problems and having distinct adaptability characteristics are applied.

\begin{figure}[!ht]
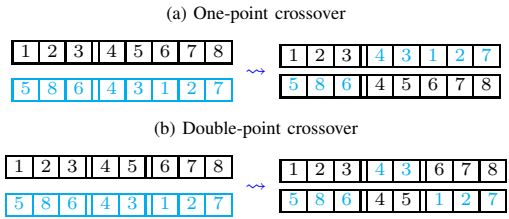

  \centering
 {\scriptsize 
\[\begin{array}{c} 
\mbox{(a) One-point crossover}\\[2mm]
\begin{array}{c}
\begin{array}{|@{\hspace{1mm}}c@{\hspace{1mm}}|@{\hspace{1mm}}c@{\hspace{1mm}}|@{\hspace{1mm}}c@{\hspace{1mm}}||@{\hspace{1mm}}c@{\hspace{1mm}}|@{\hspace{1mm}}c@{\hspace{1mm}}|@{\hspace{1mm}}c@{\hspace{1mm}}|@{\hspace{1mm}}c@{\hspace{1mm}}|@{\hspace{1mm}}c@{\hspace{1mm}}|}
\hline 1&2&3&4&5&6&7&8 \\
\hline
\end{array}\\[2mm]
{\color{cyan}\begin{array}{|@{\hspace{1mm}}c@{\hspace{1mm}}|@{\hspace{1mm}}c@{\hspace{1mm}}|@{\hspace{1mm}}c@{\hspace{1mm}}||@{\hspace{1mm}}c@{\hspace{1mm}}|@{\hspace{1mm}}c@{\hspace{1mm}}|@{\hspace{1mm}}c@{\hspace{1mm}}|@{\hspace{1mm}}c@{\hspace{1mm}}|@{\hspace{1mm}}c@{\hspace{1mm}}|}
\hline 
 5&8&6&4&3&1&2&7 \\
\hline
\end{array}}
\end{array}
{\color{blue} \huge\leadsto}
\begin{array}{c}
\begin{array}{|@{\hspace{1mm}}c@{\hspace{1mm}}|@{\hspace{1mm}}c@{\hspace{1mm}}|@{\hspace{1mm}}c@{\hspace{1mm}}||@{\hspace{1mm}}c@{\hspace{1mm}}|@{\hspace{1mm}}c@{\hspace{1mm}}|@{\hspace{1mm}}c@{\hspace{1mm}}|@{\hspace{1mm}}c@{\hspace{1mm}}|@{\hspace{1mm}}c@{\hspace{1mm}}|}
\hline 1&2&3&{\color{cyan}4}&{\color{cyan}3}&{\color{cyan}1}&{\color{cyan}2}&{\color{cyan}7} \\
\hline
\end{array}\\[1mm]
\begin{array}{|@{\hspace{1mm}}c@{\hspace{1mm}}|@{\hspace{1mm}}c@{\hspace{1mm}}|@{\hspace{1mm}}c@{\hspace{1mm}}||@{\hspace{1mm}}c@{\hspace{1mm}}|@{\hspace{1mm}}c@{\hspace{1mm}}|@{\hspace{1mm}}c@{\hspace{1mm}}|@{\hspace{1mm}}c@{\hspace{1mm}}|@{\hspace{1mm}}c@{\hspace{1mm}}|}
\hline {\color{cyan}5}&{\color{cyan}8}&{\color{cyan}6}&4&5&6&7&8 \\
\hline
\end{array}
\end{array}
\end{array}
\]
\[\begin{array}{c} 
\mbox{(b) Double-point crossover}\\[2mm]
\begin{array}{c}
\begin{array}{|@{\hspace{1mm}}c@{\hspace{1mm}}|@{\hspace{1mm}}c@{\hspace{1mm}}|@{\hspace{1mm}}c@{\hspace{1mm}}||@{\hspace{1mm}}c@{\hspace{1mm}}|@{\hspace{1mm}}c@{\hspace{1mm}}||@{\hspace{1mm}}c@{\hspace{1mm}}|@{\hspace{1mm}}c@{\hspace{1mm}}|@{\hspace{1mm}}c@{\hspace{1mm}}|}
\hline 1&2&3&4&5&6&7&8 \\
\hline
\end{array}\\[2mm]
{\color{cyan}\begin{array}{|@{\hspace{1mm}}c@{\hspace{1mm}}|@{\hspace{1mm}}c@{\hspace{1mm}}|@{\hspace{1mm}}c@{\hspace{1mm}}||@{\hspace{1mm}}c@{\hspace{1mm}}|@{\hspace{1mm}}c@{\hspace{1mm}}||@{\hspace{1mm}}c@{\hspace{1mm}}|@{\hspace{1mm}}c@{\hspace{1mm}}|@{\hspace{1mm}}c@{\hspace{1mm}}|}
\hline 
 5&8&6&4&3&1&2&7 \\
\hline
\end{array}}
\end{array}
{\color{blue} \huge\leadsto}
\begin{array}{c}
\begin{array}{|@{\hspace{1mm}}c@{\hspace{1mm}}|@{\hspace{1mm}}c@{\hspace{1mm}}|@{\hspace{1mm}}c@{\hspace{1mm}}||@{\hspace{1mm}}c@{\hspace{1mm}}|@{\hspace{1mm}}c@{\hspace{1mm}}||@{\hspace{1mm}}c@{\hspace{1mm}}|@{\hspace{1mm}}c@{\hspace{1mm}}|@{\hspace{1mm}}c@{\hspace{1mm}}|}
\hline 1&2&3&{\color{cyan}4}&{\color{cyan}3}& 6&7&8\\
\hline
\end{array}\\[1mm]
\begin{array}{|@{\hspace{1mm}}c@{\hspace{1mm}}|@{\hspace{1mm}}c@{\hspace{1mm}}|@{\hspace{1mm}}c@{\hspace{1mm}}||@{\hspace{1mm}}c@{\hspace{1mm}}|@{\hspace{1mm}}c@{\hspace{1mm}}||@{\hspace{1mm}}c@{\hspace{1mm}}|@{\hspace{1mm}}c@{\hspace{1mm}}|@{\hspace{1mm}}c@{\hspace{1mm}}|}
\hline {\color{cyan}5}&{\color{cyan}8}&{\color{cyan}6}&4&5&{\color{cyan}1}&{\color{cyan}2}&{\color{cyan}7} \\
\hline
\end{array}
\end{array}
\end{array}
\]}
\vspace{-4mm}  
  \caption{One-point and double-point crossing operators.}
  \label{fig:crossover}
\end{figure}

\begin{itemize}%[leftmargin=1mm,itemsep=0em]
    \item Simple Genetic Algorithm (\ga): 
to evolve local population, \ga\
considers a breeding cycle where the best parents are selected and produce offspring by 
applying one-point crossover (Fig.  \ref{fig:crossover} (a)). Then, the descendants replace the worst individuals in the current population. The breeding cycle relies on four parameters, namely, the percentages of {\it selection} and {\it replacement}, and the probability of application of {\it mutation} and {\it crossover}. \ga\ was developed by J. H. Holland in the 1970s \cite{Holland1973}.

    \item Double-point Crossover Genetic Algorithm (\gad): 
    it has a similar behavior than \ga\, except by the technique to promote {\it crossover}, illustrated in Fig.  \ref{fig:crossover} (b), and how the local population evolves: in contrast with \ga\, descendants replace individuals randomly selected in \gad. 
    
    \item Differential Evolution (\de):  
    it was proposed by Storn and Price \cite{DE1970}, and is a method to optimize functions over the real multidimensional space $\mathbb{R}^n$. We adapt the algorithm by restricting the domain of the function as the set of permutations. Two main parameters guide the evolutionary process: the {\it mutation factor} $F_M$, applied to individuals randomly selected from the population to generate mutants, and the {\it probability of crossover} $P_C$. 
    The local population evolves by replacing individuals having the worst fitness with mutants.

% is a optimization method for multidimensional real value functions that use a population of individual solutions, proposed by Storn and Price \cite{DE1970}.  
%  The {\it mutation factor} $F_M$, and the {\it probability of crossover} $P_C$ guide the evolutionary process. In the mutation, three distinct individuals $I_\alpha$, $I_\beta$, and $I_\gamma$ are randomly selected from the population. Then, $I_\alpha$ suffers a disturbance resulting from the vector difference between $I_\beta$ and $I_\gamma$  multiplied by $F_M$, giving rise to a new mutated individual $I_n = I_\alpha + F_M \times (I_\beta - I_\gamma)$. At each iteration, a new population is generated, replacing individuals with the worst fitness by mutants. 

    \item Self-adjusting Particle Swarm Optimization (\pso): 
    it was introduced by Eberhart and Kennedy \cite{eberhart1995particle} and is based on the behavior of social organisms in groups. Originally, \pso\ was developed to optimize continuous functions from particles' velocity and position in an $n$-dimensional space. At each iteration, the vector representing a particle velocity is built from the best positions of such a particle and all particles, the velocity of the particle in the previous iteration, the individual and global acceleration (that influence the distance a particle can cover), and the weight of inertia (momentum). In this paper, we use the \pso\ proposed in \cite{pso2011}, which is self-adaptive since momentum and the individual and global acceleration coefficients are self-tuning during the search process.

\end{itemize} 
    
%\pso\ and \de\ are known to be suitable for solving real value function problems. 

To adapt \pso\ and \de\ to URD, to each $n$-dimensional real vector $v$ randomly generated is associated a signed permutation constructed from the unsigned permutation $\pi = (\pi_1, \ldots, \pi_n)$ gave as input: if the $i-$th entry of $v$ belongs to the interval $[0, 0.5)$ then $\pi_i$ receives a negative orientation; case such a coordinate is in $[0.5, 1]$ then $\pi$ is assigned positively. However, if the continuous vector representation has an entry outside of the interval $[0,1]$, a correct orientation is randomly generated.  
For \ga\ and \gad, the orientation of the genes in each individual is randomly generated as $\pm 1$. 
After the transformation of an unsigned to signed permutations, the linear algorithm to solve the SRD problem, proposed by Bader \emph{et al.}  \cite{Bader2001linear}, computes the fitness of each particle/individual.

%  each particle/individual in the swarm/population is associated with a signed permutation randomly constructed from the unsigned permutation provided as input. For \ga\ and \gad, the orientation of the genes in each individual is randomly generated as $\pm 1$; for \pso\ and \de, the orientation is randomly generated as a value in the closed interval $[0,1]$, being values smaller than and greater than or equal to $0.5$ interpreted as negative and positive orientations, respectively.  For applying \pso\ and \de, if this continuous representation of the orientation of an individual's gene gets outside the interval $[0,1]$, a correct orientation is randomly generated.   
% The fitness of each particle/individual is computed using Bader \emph{et al.}  \cite{Bader2001linear} linear approach to solve the  SRD problem.

\section{Communication Topologies} \label{sec:topologies}
We select a static and a dynamic topology that successfully addressed URD in \cite{Lucas2020} for homogeneous PIMs and in \cite{Lucas2021} for non reconfigurable heterogeneous PIMs. We choose asynchronous models since the dynamic asynchronous HePIM was the one that provided the best results. 

\begin{figure*}[!ht]
  \centering
  \includegraphics[width=1\textwidth]{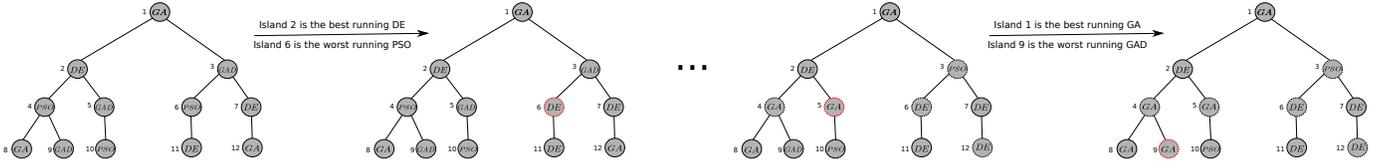}
  \caption{Example of reconfiguration in the static binary tree topology. Red dotted nodes represent islands that have undergone current reconfiguration, while black dotted nodes label islands that have undergone reconfiguration in previous cycles.}
  \label{fig:heterogeneousBTree}
\end{figure*}

\begin{figure*}[!ht]
  \centering
  \includegraphics[width=1\textwidth]{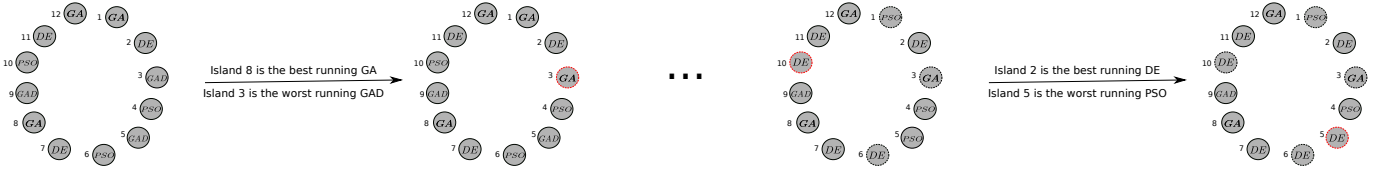}
  \caption{Example of reconfiguration in the dynamic complete graph topology. Red dotted nodes represent islands that have undergone current reconfiguration, while black dotted nodes label islands that have undergone reconfiguration in previous cycles.}
  \label{fig:heterogeneousCGraph}
\end{figure*}

 The static topology is a 12-island bi-directional binary tree (tree to the left in Figure \ref{fig:heterogeneousBTree}), and the dynamic topology is the 12-island complete graph (graph to the left in Figure \ref{fig:heterogeneousCGraph}). In the complete graph topology all pairs of islands may exchange individuals. The island communication dynamism is acquired by exploring diversity and quality into each island, given by fitness variance and average metrics. Variance measures islands' diversity: high variance represents high individuals' diversity,  improving the chances of evolution into islands. Fitness average measures the quality of island populations.
According to such metrics, the islands are ranked as {\bf good}, {\bf bad}, and {\bf medium}.  Migrations exchange individuals between good and bad islands, and medium and medium islands only (for short, {\it gbmm}).   Reconfiguration uses the same metric to update the BA executed by islands, according to the best performance experienced by other islands. So, reconfigurable HePIMs perform migration and updating of BAs at some intervals during their evolution. 

The models introduced in this paper are \rhtreeDA\ and \rhgbmmDA.  The former one uses the static tree topology and the latter one the dynamic complete graph topology. Both topologies are asynchronous and evolve through a refined migration policy that allows exchange of individuals maintaining diversity of the model, and furthermore, through the new feature of dynamic reconfiguration that allows updating the BAs executed in their islands, improving in this manner the performance of the island model.  Reconfiguration cycles for \rhtreeDA\ and \rhgbmmDA are illustrated in Figures \ref{fig:heterogeneousBTree} and \ref{fig:heterogeneousCGraph}, respectively.

\section{Experiments and analysis of accuracy}\label{sec:experimentsaccuracy}
As in \cite{Lucas2021}, all PIMs, including the new reconfigurable models were implemented using the {\tt MPI} library of {\tt C} in Linux, and for the sake of comparison, experiments were executed on a computational platform using two Xeon E5-2620 2.4 GHz six core processors with hyper-threading. 

The basis to compare the performance of PIMs, are sequential versions of \ga, \gad, \de\ and \pso\ with populations of size $24 n \log n$ and breeding cycles fixed as $n$. Also, we select eight 12-island asynchronous  HoPIMs, designed in \cite{Lucas2020}, running exclusively one of the BAs: \ga, \gad, \de\ or \pso. Furthermore, we select two asynchronous HePIMs designed in \cite{Lucas2021}.  
The homogeneous models are \gatreeDA, \gadtreeDA, \detreeDA, \psotreeDA,  \gagbmmDA, \gadgbmmDA, \degbmmDA,   and \psogbmmDA.  The superscripts denote the BA used by the homogeneous model; the subscript prefixes, whether the model use the static tree ({\tt Tr}) or the dynamic complete graph topology ({\tt gbmm}); and, the subscript suffix {\tt 12A} indicates the number of islands and that the model is asynchronous. From \cite{Lucas2021}, we select the heterogeneous PIMs \htreeDA\ and \hgbmmDA, being the latter one the HePIM that provided the best quality results.

\subsection{Parameter Setup}
  
The parameters for BAs, HoPIMs and non-reconfigurable HePIMs obtained in \cite{Lucas2021} were used. The \emph{parameter tuning} adopted the ``taxonomy T1'' in \cite{EIBEN201119}. 
Table \ref{tab:parameters} presents the parameter ranges. For percentages, the tested values range between $2\%$ and $100\%$. For probabilities, the values range from $0.02$ to $1.0$, and for the mutation parameter from $0.01$ to $0.02$. For \de, the $F_M$ parameter ranges from $1\%$ to $2\%$ since values above $2\%$ degrade the quality of solutions.
For \pso, the parameters to guide the particles in the search space are self-adjusting.

The setup tested BAs, HoPIMs and HePIMs over packages of twenty $n$-gene permutations,  $n \in \{50,60,\ldots,140,150\}$. All parameters reference values were evaluated and those that provided the best solutions  selected (see Tables \ref{table:parametersettingGA_GAD} and 
 \ref{table:parametersettingDE_PSO}).  HePIMs use the same evolutionary parameters than HoPIMs, and only migration parameters were calibrated. Reconfigurable HePIMs add reconfiguration frequency to the associated HePIMs (see Table \ref{table:parametersettingHet}).

\begin{table}[!t]
	{\small
		\caption{Estimated Values for the Parameters}
		\label{tab:parameters}
		\vspace{-3mm}
		\begin{center}
			\begin{tabular}{|c|c|c|}
				\cline{2-3}
				\multicolumn{1}{c|}{}& \multicolumn{1}{|c|}{Parameter}& \multicolumn{1}{|c|}{Estimated values}\\
				\hline
				\multirow{4}{*}{\ga\ and \gad}&\mbox{\it crossover} & $0.02, 0.04,\cdots,0.98, 1.0$ \\ \cline{2-3}
				&\mbox{\it mutation}  & $0.01, 0.011,\cdots,0.019, 0.02$\\ \cline{2-3}
			    &\mbox{\it selection} & $2\%, 4\%,\cdots,98\%, 100\%$\\ \cline{2-3}
				&\mbox{\it replacement} & $2\%, 4\%,\cdots,98\%, 100\%$ \\ \hline
				\multirow{2}{*}{\de}
				&\mbox{\it $P_C$} & $0.02, 0.04,\cdots,0.98, 1.0$ \\ \cline{2-3}
				&\mbox{\it $F_M$}  & $1\%, 1.1\%,\cdots,1.9\%, 2\%$
				\\ \hline
				\multirow{5}{*}{Migration}
				& \mbox{\it IN} & 1,2,3,4,5,6,7,8,9,10,11,12,13\\ \cline{2-3}
				&\mbox{\it EMI}  & 1=Best, 2=Worst, 3=Random \\ \cline{2-3}
				&\mbox{\it EP}  & 1=Clone, 2=Remove\\ \cline{2-3}
				&\mbox{\it IMI}  & 1=Worst, 2=Random, 3=Similar\\ \cline{2-3}
				&\mbox{\it MI} & $2\%, 4\%,\cdots,98\%, 100\%$ \\\hline
			\end{tabular}
		\end{center}}
		\vspace{-6mm}
	\end{table}

\begin{table}[!t]
        {\small
            \caption{Parameter Settings for \ga,   \gad, and associated  HoPIMs.}
            \label{table:parametersettingGA_GAD}
            \vspace{-3mm}
            \begin{center}
            \begin{tabular}
            {|@{\hspace{0.2mm}}c@{\hspace{0.2mm}}
            |@{\hspace{0.2mm}}c@{\hspace{0.2mm}}
         %   |@{\hspace{0.2mm}}c@{\hspace{0.2mm}}
            |@{\hspace{0.2mm}}c@{\hspace{0.2mm}}
         %   |@{\hspace{0.2mm}}c@{\hspace{0.2mm}}
            |@{\hspace{0.2mm}}c@{\hspace{0.2mm}}
            |@{\hspace{0.2mm}}c@{\hspace{0.2mm}}
         %   |@{\hspace{0.2mm}}c@{\hspace{0.2mm}}
            |@{\hspace{0.2mm}}c@{\hspace{0.2mm}}
         %   |@{\hspace{0.2mm}}c@{\hspace{0.2mm}}
            |@{\hspace{0.2mm}}c@{\hspace{0.2mm}}|}
                    \hline 
                \multicolumn{1}{|c|}{}&
                \multicolumn{1}{|c|}{}&
                \multicolumn{2}{|c|}{\pga}&
                \multicolumn{1}{|c|}{}&
                \multicolumn{2}{|c|}{\pgad}\\
                \hline
                    \multicolumn{1}{|@{\hspace{0.2mm}} c@{\hspace{0.2mm}}|}{Parameter} &
                    \multicolumn{1}{|@{\hspace{0.2mm}} c@{\hspace{0.2mm}}|}{\ga}&
%                    \multicolumn{1}{|@{\hspace{0.2mm}} c@{\hspace{0.2mm}}|}{\treeDS}&
                    \multicolumn{1}{|@{\hspace{0.2mm}} c@{\hspace{0.2mm}}|}{\treeDA}& 
 %                   \multicolumn{1}{|@{\hspace{0.2mm}} c@{\hspace{0.2mm}}|}{\gbmmDS}& 
                    \multicolumn{1}{|@{\hspace{0.2mm}} c@{\hspace{0.2mm}}|}{\gbmmDA}&
                    \multicolumn{1}{|@{\hspace{0.2mm}} c@{\hspace{0.2mm}}|}{\gad}&
%                    \multicolumn{1}{|@{\hspace{0.2mm}} c@{\hspace{0.2mm}}|}{\treeDS}&
                    \multicolumn{1}{|@{\hspace{0.2mm}} c@{\hspace{0.2mm}}|}{\treeDA}& 
 %                   \multicolumn{1}{|@{\hspace{0.2mm}} c@{\hspace{0.2mm}}|}{\gbmmDS}& 
                    \multicolumn{1}{|@{\hspace{0.2mm}} c@{\hspace{0.2mm}}|}{\gbmmDA}\\
                    \hline
                    {\it crossover} &$.90$   &$.98$   & $.96$ &
                    $.92$  &$.98$    & $.98$\\ \hline
                    {\it mutation}   &$.02$     &$.015$    &$.011$ &
                    $.01$     &$.01$    &$.01$ \\ \hline
                    {\it selection}    &$60\%$   &$92\%$   &$94\%$ &
                    $98\%$   &$98\%$   &$94\%$ \\ \hline
                    {\it replacement}  &$60\%$   &$70\%$  &$70\%$  &
                    $90\%$   &$80\%$  &$90\%$ \\ \hline
                    \mbox{\it IN}  &  &9  &5 &  &12  &5 \\ \hline
                    \mbox{\it EMI} &  &1  &1 &  &1  &1 \\ \hline
                    \mbox{\it EP}  &  &2  &2 &  &2  &1\\ \hline
                    \mbox{\it IMI} &  &1  &1 &  &1  &1 \\ \hline
                    \mbox{\it MI}  &  &$30\%$  &$30\%$&                  &$14\%$  &$12\%$ \\\hline
                \end{tabular}
          \end{center}
}
            \vspace{-5mm}
        \end{table}

\begin{table}[!t]
        {\small
            \caption{Parameter Settings for \de, \pso, and associated HoPIMs.}
            \label{table:parametersettingDE_PSO}
            \vspace{-3mm}
            \begin{center}
            \begin{tabular} 
            {|@{\hspace{0.2mm}}c@{\hspace{0.2mm}}
            |@{\hspace{0.2mm}}c@{\hspace{0.2mm}}
        %    |@{\hspace{0.2mm}}c@{\hspace{0.2mm}}
            |@{\hspace{0.2mm}}c@{\hspace{0.2mm}}
        %    |@{\hspace{0.2mm}}c@{\hspace{0.2mm}}
            |@{\hspace{0.2mm}}c@{\hspace{0.2mm}}
        %    |@{\hspace{0.2mm}}c@{\hspace{0.2mm}}
            |@{\hspace{0.2mm}}c@{\hspace{0.2mm}}
        %    |@{\hspace{0.2mm}}c@{\hspace{0.2mm}}
            |@{\hspace{0.2mm}}c@{\hspace{0.2mm}}|}
            \hline
            \multicolumn{1}{|c|}{} &
            \multicolumn{1}{|c|}{} &
             \multicolumn{2}{|c|}{\pde} &
            \multicolumn{2}{|c|}{\ppso}
            \\
                    \hline
                    \multicolumn{1}{|@{\hspace{0.2mm}} c@{\hspace{0.2mm}}|}{Parameter}& 
                    \multicolumn{1}{|@{\hspace{0.2mm}} c@{\hspace{0.2mm}}|}{\de}&
                %    \multicolumn{1}{|@{\hspace{0.2mm}} c@{\hspace{0.2mm}}|}{\treeDS}&
                    \multicolumn{1}{|@{\hspace{0.2mm}} c@{\hspace{0.2mm}}|}{\treeDA}& 
                 %   \multicolumn{1}{|@{\hspace{0.2mm}} c@{\hspace{0.2mm}}|}{\gbmmDS}& 
                    \multicolumn{1}{|@{\hspace{0.2mm}} c@{\hspace{0.2mm}}|}{\gbmmDA}&
                 %   \multicolumn{1}{|@{\hspace{0.2mm}} c@{\hspace{0.2mm}}|}{\treeDS} &
                    \multicolumn{1}{|@{\hspace{0.2mm}} c@{\hspace{0.2mm}}|}{\treeDA}& 
                 %   \multicolumn{1}{|@{\hspace{0.2mm}} c@{\hspace{0.2mm}}|}{\gbmmDS}& 
                    \multicolumn{1}{|@{\hspace{0.2mm}} c@{\hspace{0.2mm}}|}{\gbmmDA}\\
                    \hline
                    $P_C$          &$.74$    &$.72$    &$.78$ &&\\ \hline
                    $F_M$          & $1\%$      &$1.4\%$  &$1\%$ &&\\ \hline
                    \mbox{\it IN}  &  &3  &5 &6  &5\\ \hline
                    \mbox{\it EMI} &  &1  &1 &3 &3\\ \hline
                    \mbox{\it EP}  &  &1   &2  &2  &2\\ \hline
                    \mbox{\it IMI} &  &1  &1  &1  &2   \\ \hline
                    \mbox{\it MI}  &  &$14\%$  &$12\%$  &$12\%$ &$22\%$\\ \hline
                \end{tabular}
            \end{center}}
            \vspace{-5mm}
        \end{table}
\begin{table}[!t]
        {\small
            \caption{Parameter Settings for HePIMs.}
            \label{table:parametersettingHet}
            \vspace{-3mm}
            \begin{center}
            \begin{tabular}
            {|@{\hspace{0.2mm}}c@{\hspace{0.2mm}}
            |@{\hspace{0.2mm}}c@{\hspace{0.2mm}}
            |@{\hspace{0.2mm}}c@{\hspace{0.2mm}}
            |@{\hspace{0.2mm}}c@{\hspace{0.2mm}}
            |@{\hspace{0.2mm}}c@{\hspace{0.2mm}}
            |@{\hspace{0.2mm}}c@{\hspace{0.2mm}}
            |@{\hspace{0.2mm}}c@{\hspace{0.2mm}}
            |@{\hspace{0.2mm}}c@{\hspace{0.2mm}}
            |@{\hspace{0.2mm}}c@{\hspace{0.2mm}}|}
                    \hline
                    \multicolumn{1}{|@{\hspace{0.2mm}} c@{\hspace{0.2mm}}|}{Parameter}& 
                    \multicolumn{1}{|@{\hspace{0.2mm}} c@{\hspace{0.2mm}}|}{\htreeDA}&
                    \multicolumn{1}{|@{\hspace{0.2mm}} c@{\hspace{0.2mm}}|}{\rhtreeDA}& 
                    \multicolumn{1}{|@{\hspace{0.2mm}} c@{\hspace{0.2mm}}|}{\hgbmmDS}& 
                    \multicolumn{1}{|@{\hspace{0.2mm}} c@{\hspace{0.2mm}}|}{\rhgbmmDA}\\
                    \hline
                    \mbox{\it IN}  &3 &3 &6 &6 \\ \hline
                    \mbox{\it EMI} &1 &1 &3 &3 \\ \hline
                    \mbox{\it EP}  &2 &2 &1 &1 \\ \hline
                    \mbox{\it IMI} &3 &3 &3 &3\\ \hline
                    \mbox{\it MI} &$10\%$ &$10\%$ & $14\%$ & $14\%$ \\\hline
                    \mbox{\it RF}  & & 14\% & &24\% \\ \hline
                \end{tabular}
            \end{center}}
            \vspace{-8mm}
        \end{table}

\subsection{Analysis of Accuracy}\label{sec:analysis}

HePIMs use parameters taken from Tables \ref{table:parametersettingGA_GAD}, \ref{table:parametersettingDE_PSO} and \ref{table:parametersettingHet} according to the parameter setting obtained in \cite{Lucas2021}. In addition, the new reconfigurable HePIMs performed reconfigurations after each 14\% and 24\% generations, giving a total of seven and four reconfiguration cycles,  respectively, for the static tree and dynamic complete graph topologies.

For each permutation size, $n \in \{100,110,\ldots,150\}$, one package of one hundred unsigned permutations with $n$ genes was randomly generated.   
All PIMs were executed ten times using each one of the $n$ permutations of size $n$, and the average of these executions for each permutation is taken as the result. The average gives the computed number of reversals for each unsigned permutation. 

\begin{figure}[!ht]
\centering
\includegraphics[width=0.46\textwidth]{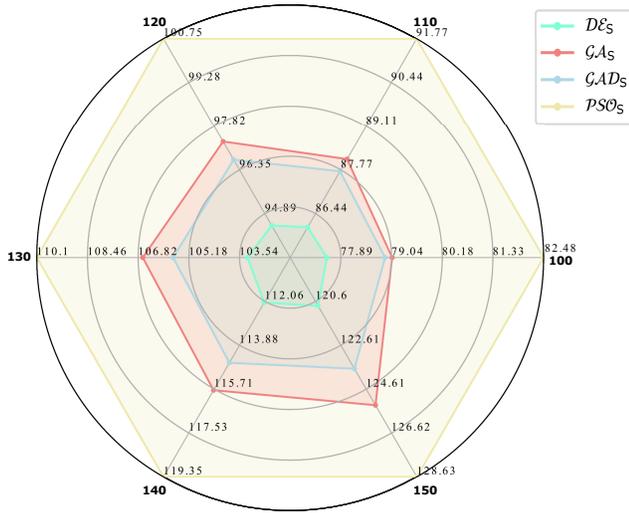}
  \caption{Accuracy of the sequential BAs \de, \ga, \gad\ and \pso.}
  \label{fig:sequentialPIMs}
\end{figure}

The accuracies of non- and reconfigurable HePIMs are compared. The radar chart in Fig. \ref{fig:sequentialPIMs}, from previous experiments, shows that \de\ is the best adapted BA for the URD problem \cite{Lucas2021}.  In contrast,  \pso\ provided the worst results, while   \ga\ and \gad, in this order, gave competitive results.   The six radii of the chart represent the accuracy for inputs of size $100, 110$ to $150$. The ranges and scales in each radius of the radar chart are adjusted for the sake of presentation.

\begin{figure*}[!ht]
\centering
\includegraphics[width=0.52\textwidth]{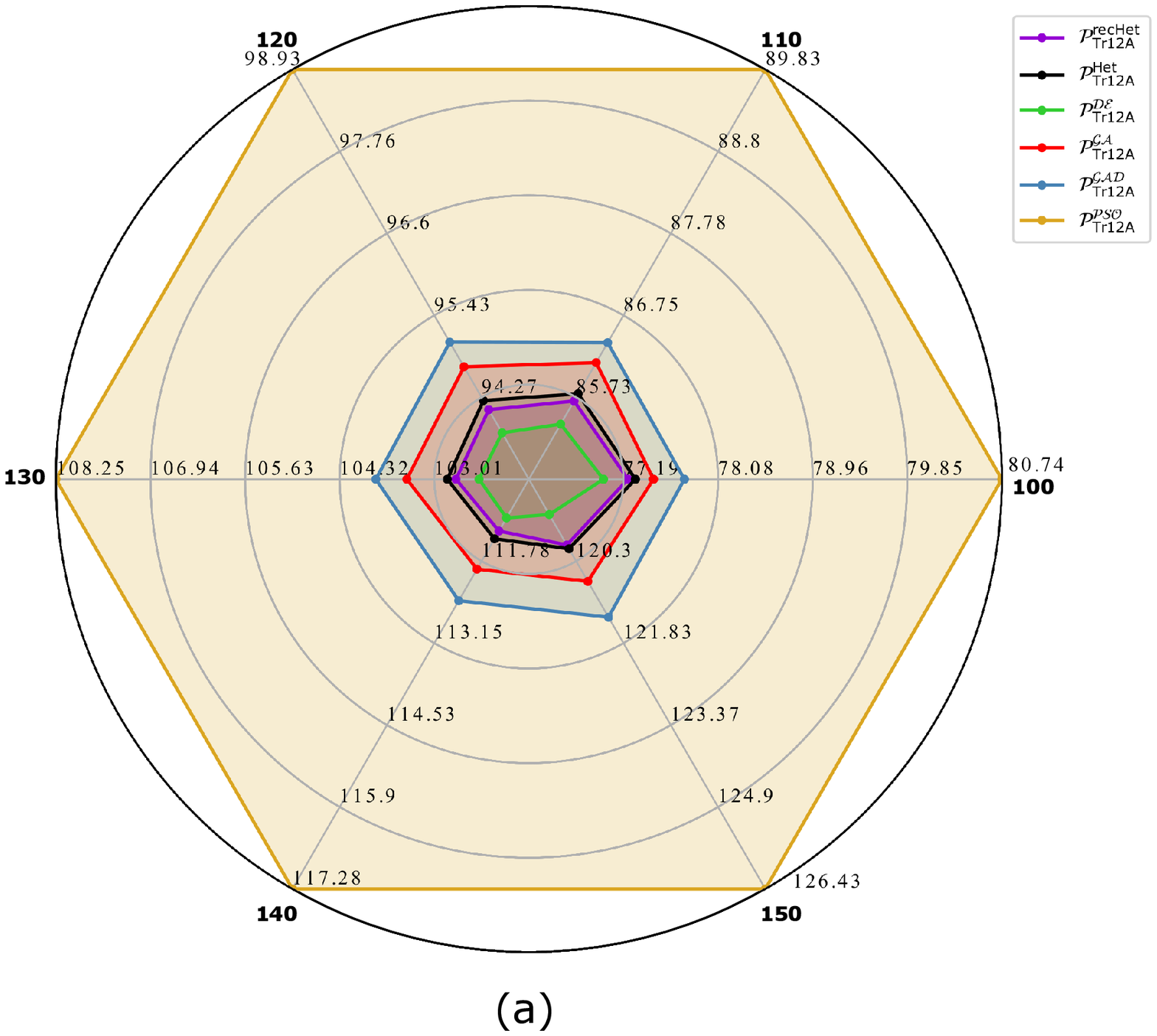}\hspace{-2mm}
  \includegraphics[width=0.48\textwidth]{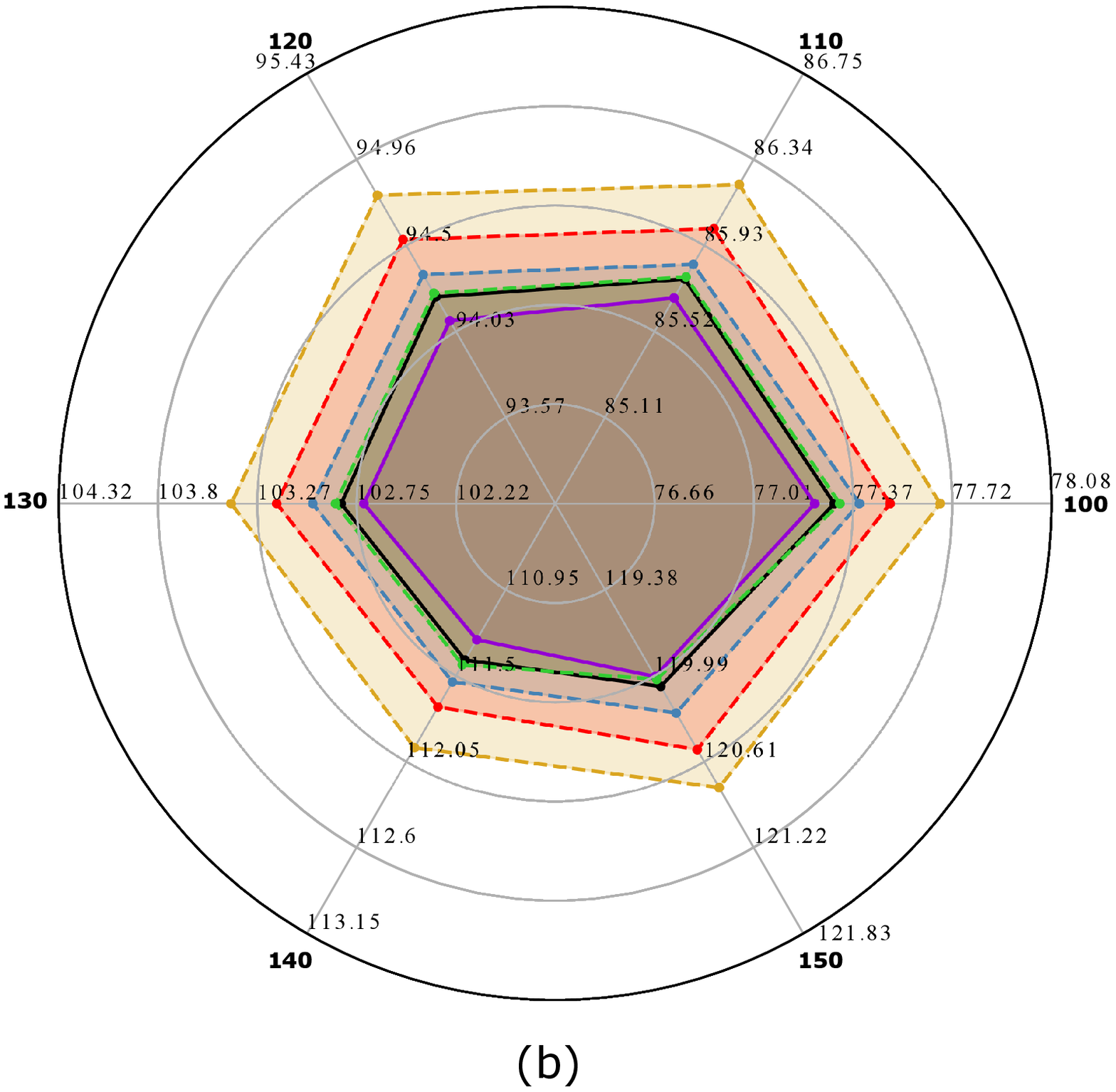}
 \caption{(a) Accuracy of \rhtreeDA, \htreeDA and related HoPIMs; (b) Accuracy of \rhtreeDA, \htreeDA and  average results of each set of  islands in \htreeDA\ running the same BA.}
  \label{fig:het_tree12A}
    \vspace{-4mm}
\end{figure*}

\begin{figure*}[!ht]
\centering
\includegraphics[width=0.528\textwidth]{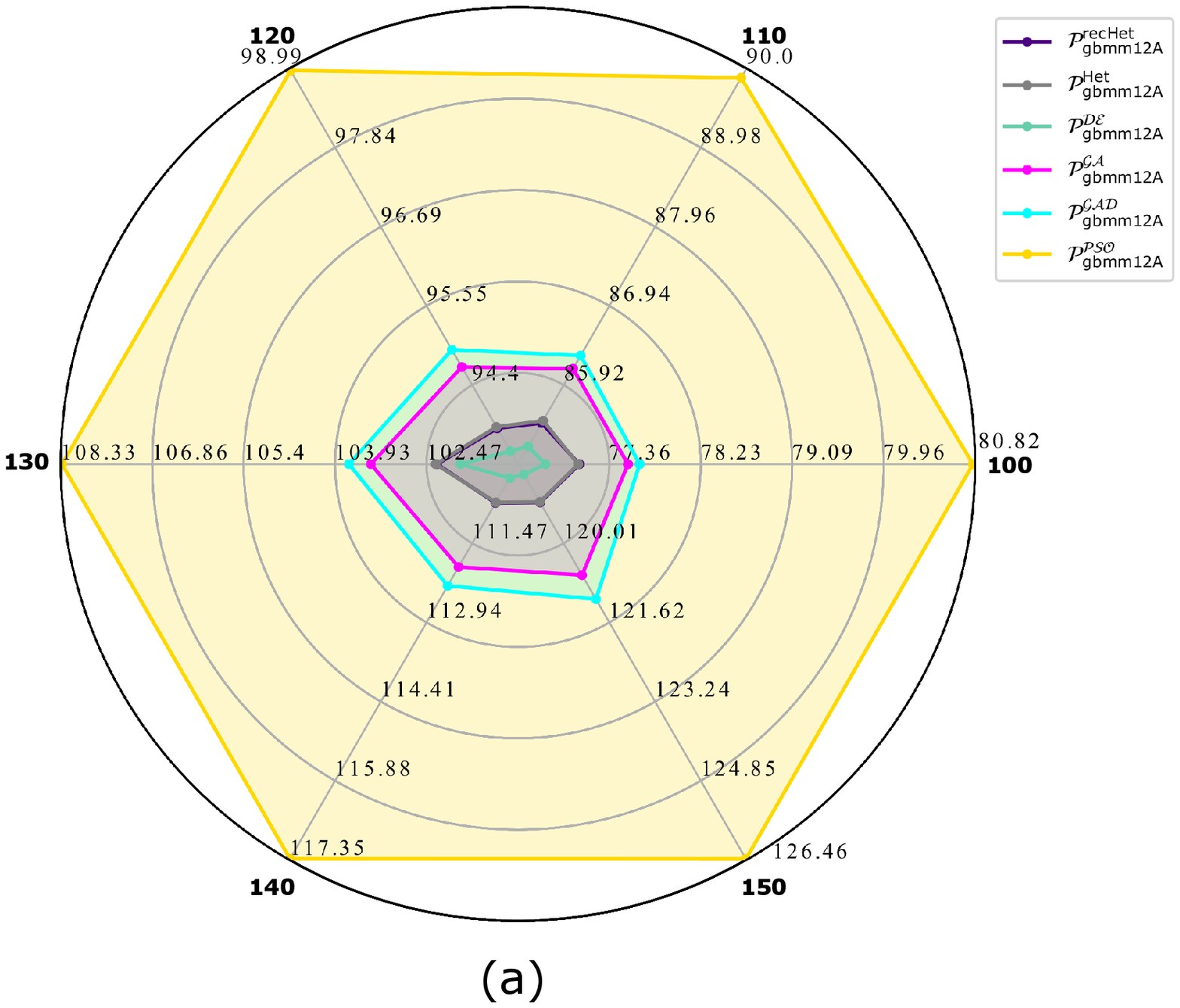}\hspace{-2mm}
  \includegraphics[width=0.462\textwidth]{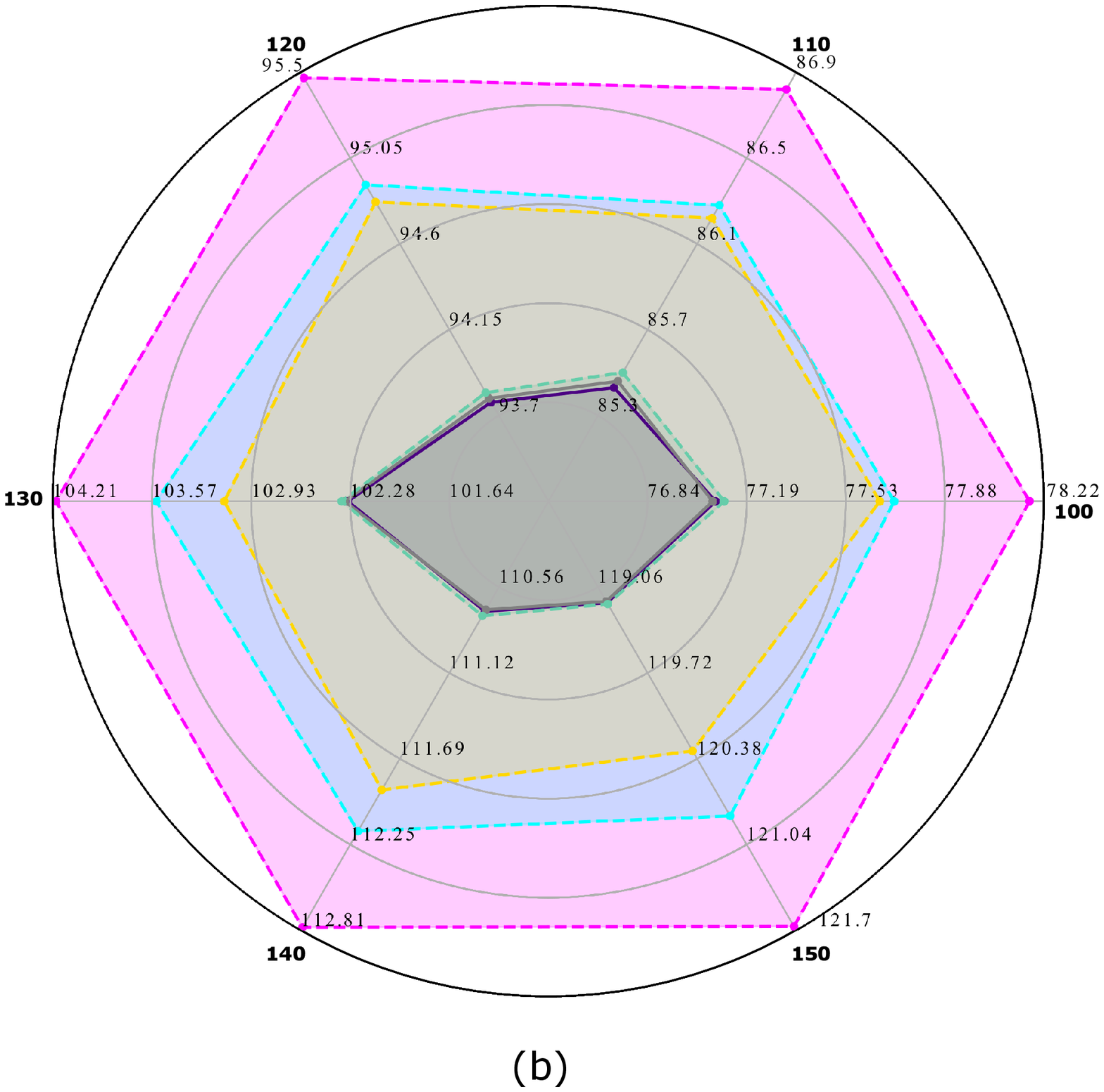}
 \caption{(a) Accuracy of \rhgbmmDA, \hgbmmDA and  related HoPIMs; (b) Accuracy of \rhgbmmDA, \hgbmmDA and  average results of each set of  islands in \hgbmmDA\ running the same BA.}
  \label{fig:het_gbmm12A}
    \vspace{-4mm}
\end{figure*}

PIMs with the tree and the complete graph topologies outperformed, as expected, their sequential versions \cite{Lucas2021}. The radar charts to the left in Figs. \ref{fig:het_tree12A} and \ref{fig:het_gbmm12A} show that the HoPIMs maintained the order of competitivity of their best and worst adapted BAs: the best quality solutions were obtained by \detreeDA and the worst by \psotreeDA\, for the static tree model, while for the dynamic complete graph topology, the best solutions were computed by \degbmmDA\ and the worst by \psogbmmDA.  In contrast to the fact that \gad\ provided better accuracy than \ga,  the homogeneous models \gatreeDA\ and \gagbmmDA\ respectively outperformed \gadtreeDA\ and \gadgbmmDA.

Table \ref{tab:reconfigEnd} exemplify the final island configuration. We ran \rhtreeDA and \rhgbmmDA\ over one hundred entries of size 100 and computed the average of the final distribution of the four BAs over the islands. Surprisingly the proportion of islands running \ga\ is dominant for \rhtreeDA\ that can be explained since the final average results of sets of islands with the same BA are very similar for this model (see right chart on Fig. \ref{fig:het_tree12A}). On the other side, the distribution BAs over islands for \rhgbmmDA\ is better balanced (cf. right chart on Fig. \ref{fig:het_gbmm12A}).

%MAR: Lucas, troquei o texto. 
% \lucas{
% Table \ref{tab:reconfigEnd} presents an analysis of the islands of the reconfigurable heterogeneous PIMS after completing the breeding cycle. The analysis takes one hundred entries of size 100, where we verify the percentage of algorithms used as evolutionary engine in the set of 12 islands. Regarding the static models, we have the genetic algorithm being dominant in approximately 50\% of the total islands, as seen in \cite{Lucas2021}, the set of islands with the \ga\ manages to achieve the second best result being surpassed only by the set executing \de, thus way, driving the improvement of the reconfigurable heterogeneous PIM when compared to the non-reconfigurable. On the other hand, we have a balanced scenario for the dynamic reconfigurable heterogeneous PIM, however, the results did not show significant improvements compared to the non-reconfigurable and for many inputs as seen in Fig. \ref{fig:heterogeneousPIMs}, very closed results.
% }

Figs. \ref{fig:het_tree12A} and \ref{fig:het_gbmm12A} include the accuracy for the HePIMs \htreeDA\ and \hgbmmDA, respectively.  The charts on the left show that the accuracy of these models is very competitive regarding the HoPIMs with the same architecture being only surpassed by \detreeDA\ and \degbmmDA, respectively.  The  charts on the right show (dashed lines) the best final average results obtained by each set of three islands in the HePIMs (\htreeDA, and \hgbmmDA, respectively) that execute the same BA. 
These charts make it evident that the migration policy and application of diverse BAs in the islands of the heterogeneous architectures successfully propagates the results obtained in all islands, but are not enough to outperform the quality of the homogeneous model running the best adapted BA, \detreeDA\ and \degbmmDA, respectively. 

Performance of the new reconfigurable models is also included in Figs. \ref{fig:het_tree12A} and \ref{fig:het_gbmm12A}.   The reconfigurable HePIM \rhtreeDA\ computed better quality results than the pure heterogeneous model \htreeDA, and closed the gap between this and the best adapted homogeneous architecture \detreeDA, running DE (see radar chart on the left in Fig. \ref{fig:het_tree12A}). It makes evident that adding the versatility of reconfiguration to the heterogeneous PIMs may improve their performance.   On the other side, the reconfigurable HePIM \rhgbmmDA\ computed quality results that are indistinguishable from the competitive ones computed by the non reconfigurable heterogeneous model, \hgbmmDA.

\begin{figure}[!ht]
\centering
\includegraphics[width=0.48\textwidth]{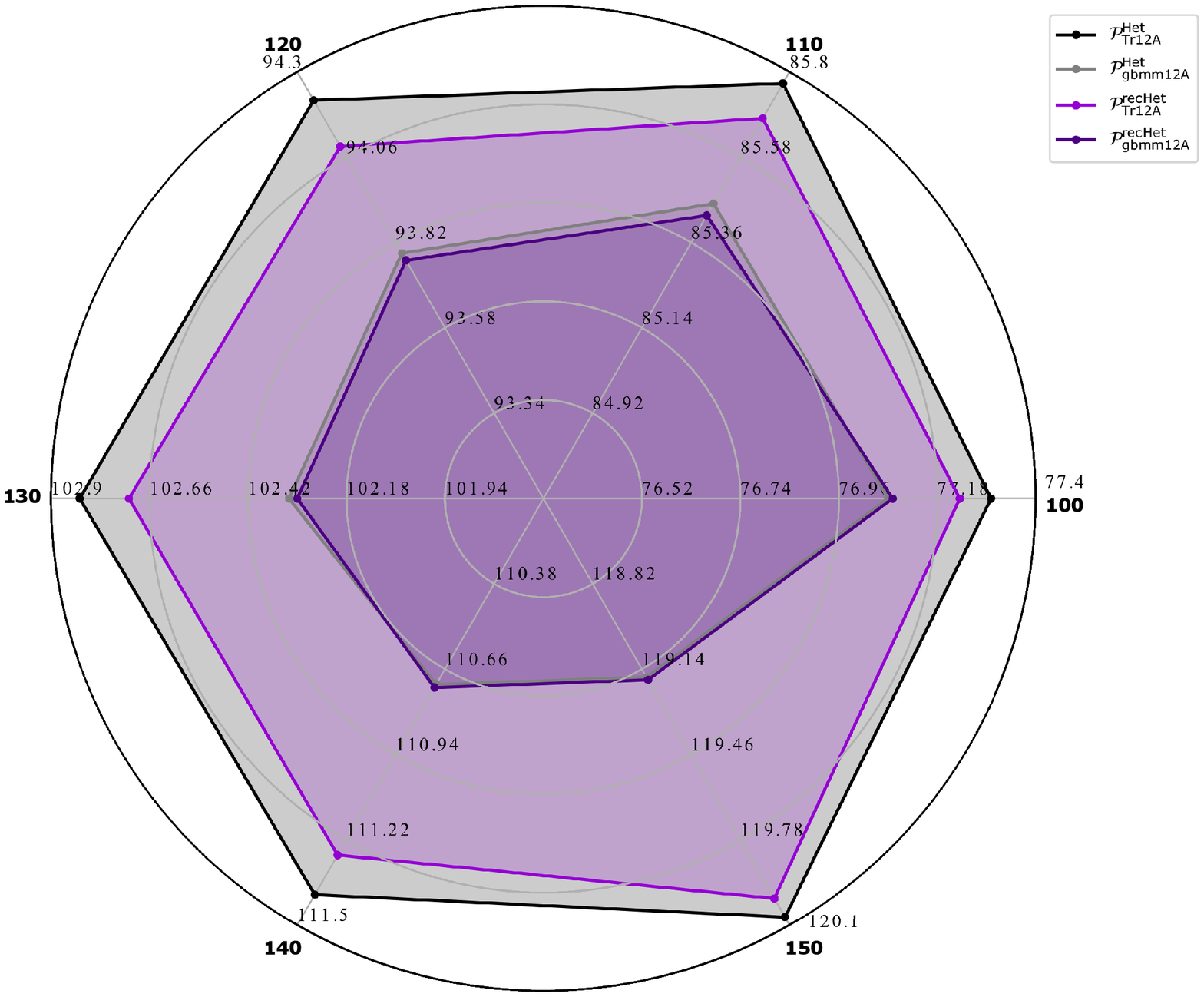}
  \caption{Accuracy of the non reconfigurable and reconfigurable HePIMs}
  \label{fig:heterogeneousPIMs}
  %  \vspace{-4mm}
\end{figure}

Fig. \ref{fig:heterogeneousPIMs} compares the accuracy of the four non- and reconfigurable HePIMs: \htreeDA, \rhtreeDA, \rhgbmmDA, and. \hgbmmDA. 
From the experiments, it is clear that the new reconfigurable heterogeneous architectures, \rhtreeDA\ and \rhgbmmDA, add to the versatility of heterogeneous PIMs the flexibility of dynamically updating the BA executed in each island promoting in this manner not only data diversity, but also algorithmic dynamism.  Reconfigurable heterogeneous PIMs open up a promising new exploration space where, unlike von Newman's style of algorithmic exploration focused on efficient information data management, algorithmic data dynamism enters as a crucial player in the game (see, for example, \cite{Hartenstein2010}, \cite{Hartenstein2013}).

\begin{table}[!t]
	{\small
		\caption{Example of the final distribution of \ga, \gad, \de\ and \pso.}
		\label{tab:reconfigEnd}
		\begin{center}
			\begin{tabular}{|c|c|c|c|c|}
				\cline{2-5}
				\multicolumn{1}{c|}{}& 
				\multicolumn{1}{|c|}{\ga}& \multicolumn{1}{|c|}{\gad}&
				\multicolumn{1}{|c|}{\de} &
				\multicolumn{1}{|c|}{\pso}
				\\
				\hline
				\multirow{1}{*}{\rhtreeDA}& 49.25\%& 4.09\% &  18.08\% & 28.58\% \\[1mm]  \hline
				\multirow{1}{*}{\rhgbmmDA}
				&29.42\% &17.5\% & 28.25\% & 24.83\% \\[1mm] \hline
			\end{tabular}
		\end{center}}
	%	\vspace{-4mm}
	\end{table}

\subsection{Statistical Analysis}\label{ssec:statisticalanalysis}

Statistical tests validated experiments using $95\%$  of significance level represented in the tests as $\alpha = 0.05$. 
The samples are the sets of one hundred outputs obtained in Section \ref{sec:analysis}.
Initially, the Friedman test was applied to define the control algorithm. Then, Holm's test was applied to check the null hypothesis that the performance of the control algorithm is the same concerning the remaining algorithms, according to Garc\'ia and Herrera approach \cite{garcia2008} (see also \cite{demvsar2006statistical}, and \cite{derrac2011}).

Table \ref{table:holmStaticHe} presents the statistical results for the PIMs with binary tree topology: \detreeDA, \htreeDA, and \rhtreeDA.  The Friedman test selects \detreeDA\ as the control algorithm, and the null hypothesis is discarded for all samples. The model \rhtreeDA\  is the second-best confirming the discussion in Section \ref{sec:analysis}. Table \ref{table:holmDynamicHe} shows the statistical results for the models with complete graph topology, \hgbmmDA, \rhgbmmDA, and the best homogeneous version, \degbmmDA, which is selected as the control algorithm. Finally, Table \ref{table:holmHeterogeneous} gives the statistical tests for the four reconfigurable and non-reconfigurable HePIMs.  The selected control algorithm was \hgbmmDA. Holm's procedure rejects the null hypotheses (for $p$-value $\le0.05$), thus, \hgbmmDA\ has statistical significance only for \htreeDA\ and \rhtreeDA, and compared to \rhgbmmDA\ there is no statistical significance confirming the discussion in Section \ref{sec:analysis}.

\begin{table}[ht]
  { \scriptsize
  \caption{Holm test for the tree topology PIMs. }
  \label{table:holmStaticHe}
  \begin{tabular}{|c|c|c|c|c|c|}
    \hline
    \multicolumn{1}{|c|}{\bfseries L}
    & \multicolumn{1}{|c|}{\bfseries Control}
    & \multicolumn{1}{|c|}{\bfseries i}
    & \multicolumn{1}{|c|}{\bfseries Algorithm}
    & \multicolumn{1}{|c|}{\bfseries \boldmath{ $p$}-value}
    & \multicolumn{1}{|c|}{\bfseries \boldmath{\!$\alpha / i$\!}}\\
    \multicolumn{1}{|c|}{\bfseries }
    & \multicolumn{1}{|c|}{\bfseries Algorithm}
    & \multicolumn{1}{|c|}{\bfseries }
    & \multicolumn{1}{|c|}{\bfseries }
    %& \multicolumn{1}{|c|}{\bfseries }
    & \multicolumn{1}{|c|}{\bfseries }
    & \multicolumn{1}{|c|}{\bfseries }\\[0.5mm]
    \hline
    \multirow{3}{*}{100} 
    & \multirow{3}{*}{}
     &  2 & \htreeDA  & 7.74959960741012E-11 & 0.025 \\[.26mm] \cline{3-6}
    & \detreeDA & 1 & \rhtreeDA  & 1.8505741383968991E-9 & 0.050 \\[0.5mm]   \hline
    \multirow{3}{*}{110}
    & \multirow{3}{*}{}
     & 2 & \htreeDA & 1.346035421050821E-15 & 0.025 \\[.26mm] \cline{3-6}
    & \detreeDA & 1 & \rhtreeDA  & 5.560820039745642E-15 & 0.050 \\[0.5mm]
    \hline
     \multirow{3}{*}{120}
    & \multirow{3}{*}{}
      & 2 & \htreeDA  & 3.722840351917189E-19 & 0.025 \\[.26mm] \cline{3-6}
    & \detreeDA & 1 & \rhtreeDA  & 3.2599374788722365E-13 & 0.050  \\[0.5mm] 
    \hline
    \multirow{3}{*}{130}
    & \multirow{3}{*}{}
     & 2 & \htreeDA  & 4.218936534105464E-12 & 0.025 \\[.26mm] \cline{3-6}
    & \detreeDA & 1 & \rhtreeDA  & 1.449490502746956E-11 & 0.050  \\[0.5mm] 
    \hline
    \multirow{3}{*}{140}
    & \multirow{3}{*}{}
    &    2 & \htreeDA    & 3.1593469401304723E-16 & 0.025 \\[.26mm] \cline{3-6}
    & \detreeDA & 1 & \rhtreeDA  & 1.4870457587052685E-9 & 0.050 \\[0.5mm] 
    \hline
    \multirow{3}{*}{150}
    & \multirow{3}{*}{}
     & 2 & \htreeDA  & 5.124221656690746E-19 & 0.025 \\[.26mm] \cline{3-6}
    &  \detreeDA& 1 & \rhtreeDA  & 9.723622409009922E-15 & 0.050  \\[0.5mm] 
    \hline
  \end{tabular} 
  \vspace{-2mm}}
\end{table}

\begin{table}[ht]
  { \scriptsize
  \caption{Holm test for the complete graph PIMs. }
  \label{table:holmDynamicHe}
  \begin{tabular}{|c|c|c|c|c|c|}
    \hline
    \multicolumn{1}{|c|}{\bfseries L}
    & \multicolumn{1}{|c|}{\bfseries Control}
    & \multicolumn{1}{|c|}{\bfseries i}
    & \multicolumn{1}{|c|}{\bfseries Algorithm}
    & \multicolumn{1}{|c|}{\bfseries \boldmath{ $p$}-value}
    & \multicolumn{1}{|c|}{\bfseries \boldmath{\!$\alpha / i$\!}}\\
    \multicolumn{1}{|c|}{\bfseries }
    & \multicolumn{1}{|c|}{\bfseries Algorithm}
    & \multicolumn{1}{|c|}{\bfseries }
    & \multicolumn{1}{|c|}{\bfseries }
    %& \multicolumn{1}{|c|}{\bfseries }
    & \multicolumn{1}{|c|}{\bfseries }
    & \multicolumn{1}{|c|}{\bfseries }\\[0.5mm]
    \hline
    \multirow{3}{*}{100} 
    & \multirow{3}{*}{}
     & 2 &  \rhgbmmDA  & 7.43098372370352E-7 & 0.025 \\[.26mm] \cline{3-6}
    & \degbmmDA & 1 & \hgbmmDA  & 1.1648657367238803E-5 & 0.050 \\[0.5mm]   \hline
    \multirow{3}{*}{110}
    & \multirow{3}{*}{}
     & 2 &  \rhgbmmDA & 9.672204071723814E-19 & 0.025 \\[.26mm] \cline{3-6}
    &  \degbmmDA& 1 & \hgbmmDA  & 2.9294885290101255E-14 & 0.050  \\[0.5mm] 
    \hline
     \multirow{3}{*}{120}
    & \multirow{3}{*}{}
     & 2 &  \hgbmmDA  & 1.792019989925749E-15 & 0.025 \\[.26mm] \cline{3-6}
    & \degbmmDA & 1 & \rhgbmmDA  & 1.792019989925749E-15 & 0.050  \\[0.5mm] 
    \hline
    \multirow{3}{*}{130}
    & \multirow{3}{*}{}
     & 2 &  \rhgbmmDA  & 3.943363947351002E-17 & 0.025 \\[.26mm] \cline{3-6}
    &  \degbmmDA & 1 & \hgbmmDA  & 2.3828362635579084E-15 & 0.050  \\[0.5mm] 
    \hline
    \multirow{3}{*}{140}
    & \multirow{3}{*}{}
     & 2 &  \rhgbmmDA  & 4.82026808703977E-22 & 0.025 \\[.26mm] \cline{3-6}
    & \degbmmDA & 1 & \hgbmmDA  & 1.0213251630273183E-20 & 0.050 \\[0.5mm] 
    \hline
    \multirow{3}{*}{150}
    & \multirow{3}{*}{}
     & 2 &  \hgbmmDA & 3.4176814448375205E-24 & 0.025 \\[.26mm] \cline{3-6}
    & \degbmmDA & 1 & \rhgbmmDA  & 1.440728401105864E-23 & 0.050  \\[0.5mm] 
    \hline
  \end{tabular} 
  \vspace{-2mm}}
\end{table}
\begin{table}[ht]
  { \scriptsize
  \caption{Holm test for reconf. and non-reconf. HePIMs. }
  \label{table:holmHeterogeneous}
  \begin{tabular}{|c|c|c|c|c|c|}
    \hline
    \multicolumn{1}{|c|}{\bfseries L}
    & \multicolumn{1}{|c|}{\bfseries Control}
    & \multicolumn{1}{|c|}{\bfseries i}
    & \multicolumn{1}{|c|}{\bfseries Algorithm}
    & \multicolumn{1}{|c|}{\bfseries \boldmath{ $p$}-value}
    & \multicolumn{1}{|c|}{\bfseries \boldmath{\!$\alpha / i$\!}}\\
    \multicolumn{1}{|c|}{\bfseries }
    & \multicolumn{1}{|c|}{\bfseries Algorithm}
    & \multicolumn{1}{|c|}{\bfseries }
    & \multicolumn{1}{|c|}{\bfseries }
    %& \multicolumn{1}{|c|}{\bfseries }
    & \multicolumn{1}{|c|}{\bfseries }
    & \multicolumn{1}{|c|}{\bfseries }\\[0.5mm]
    \hline
    \multirow{3}{*}{100} 
    & \multirow{3}{*}{}
     & 3 &  \htreeDA  & 6.219448336201955E-7 & 0.016 \\[.26mm] \cline{3-6}
    & \hgbmmDA & 2 & \rhtreeDA  & 3.5448303585580045E-5 & 0.025 \\[.26mm] \cline{3-6}
    & & 1 &  \rhgbmmDA  & 0.3958980057181269 & 0.05\\[0.5mm]   \hline
    \multirow{3}{*}{110}
    & \multirow{3}{*}{}
     & 3 &  \htreeDA  & 2.1442166253671064E-13 & 0.016 \\[.26mm] \cline{3-6}
    & \hgbmmDA & 2 & \rhtreeDA  & 7.221976514824151E-13 & 0.025 \\[.26mm] \cline{3-6}
    & & 1 &  \rhgbmmDA  & 0.1709035202307971 & 0.05\\[0.5mm]   \hline
     \multirow{3}{*}{120}
    & \multirow{3}{*}{}
      & 3 &  \htreeDA  & 2.1442166253672077E-13 & 0.016 \\[.26mm] \cline{3-6}
    & \hgbmmDA & 2 & \rhtreeDA  & 3.908557567773197E-9 & 0.025 \\[.26mm] \cline{3-6}
    & & 1 &  \rhgbmmDA  & 0.7218258402177081 & 0.05\\[0.5mm]   \hline 
    \multirow{3}{*}{130}
    & \multirow{3}{*}{}
     & 3 &  \htreeDA  & 1.076195601000617E-12 & 0.016 \\[.26mm] \cline{3-6}
    & \hgbmmDA & 2 & \rhtreeDA  & 3.415515804642443E-11 & 0.025 \\[.26mm] \cline{3-6}
    & & 1 &  \rhgbmmDA  & 0.4113137917762579 & 0.05\\[0.5mm]   \hline
    \multirow{3}{*}{140}
    & \multirow{3}{*}{}
      & 3 &  \htreeDA  & 4.342593992240847E-19 & 0.016 \\[.26mm] \cline{3-6}
    & \hgbmmDA & 2 & \rhtreeDA  & 2.1442166253671531E-13 & 0.025 \\[.26mm] \cline{3-6}
    & & 1 &  \rhgbmmDA  & 0.8694817827381613 & 0.05\\[0.5mm]   \hline
    \multirow{3}{*}{150}
    & \multirow{3}{*}{}
      & 3 &  \htreeDA  & 3.3892419952526653E-19 & 0.016 \\[.26mm] \cline{3-6}
    & \hgbmmDA & 2 & \rhtreeDA  & 2.4806730968270847E-15 & 0.025 \\[.26mm] \cline{3-6}
    & & 1 &  \rhgbmmDA  & 0.912770619096563 & 0.05\\[0.5mm]   \hline
  \end{tabular} 
  \vspace{-2mm}}
\end{table}

% and present statistical significance for most analyzed algorithms. 

% (see in Tables \ref{table:holmP1} and \ref{table:holmP2}) and the $p$-value in bold for size 140 samples indicate that the control algorithm does not have statistically significance difference regarding the algorithm in the respective row.

\section{Related Work}\label{sec:relatedwork}

As far as we know, no HePIM has been proposed that may dynamically update their BAs as proposed in this work. Here, we discuss a few works related to non-reconfigurable HePIMs.  

Bianchini and Brown \cite{Bianchini1993} proposed HePIMs with ring and torus topologies and applied them to the task map scheduling problem showing that HePIMs compute better solutions than HoPIMs. In addition, they observed that adding islands is better than increasing the population.
Also, Lin \textit{et al.} \cite{Shyn1994} proposed HePIMs, considering several migration strategies and topologies addressing the graph partitioning problem. They showed that 25-island PIMs are better than the sequential GA,  using a migration strategy that replaces the worst individuals on target islands. Furthermore,  they showed that exchanging individuals regarding their fitness-based population similarity instead gets good results without speed degradation.

Izzo \textit{et al.} \cite{sinc2009} 
proposed an asynchronous-migration HePIM from variations of the \de\ algorithm. Asynchrony was shown more intuitive and suitable over TCP/IP, where resources might become available/unavailable at any time. Izzo \textit{et al.} models showed better performance than their sequential versions. 

Gong and  Fukunaga \cite{GoFu2011} proposed a \ga-based HePIM that randomly selects different parameters for each processor. Some processors are expected to be assigned parameters that perform well on a given problem.  
Such model may be considered a one-cycle reconfigurable model. However, it applies only an initial adjust of the same algorithm and does not update BAs dynamically as ours reconfigurable HePIMs. 

Duarte \textit{et al.} \cite{duarte2018}  proposed an  attractiveness-based migration policy for five-island HePIMs that is based on island solutions' quality. Attractiveness was adjusted in \cite{Duarte2020}, inspired by the natural phenomenon known as stigmergy \cite{Capriles2007}, and the mechanism to compute islands' connections.

Silveira {\em et al.} \cite{lucas2016} proposed HoPIMs for a sequential \ga\ introduced in \cite{lucas2015} to solve the unsigned translocation problem.  Such PIMs outperformed the accuracy obtained by the \ga\ after careful calibration of the migration and breeding cycle parameters and exploration of a variety of topologies \cite{silveira2018,silveira2019}. Further,  Silveira {\em et al.} \cite{Lucas2020} analyzed synchronous HoPIMs for \ga,  \pso, and the Social Spider Algorithm (SSA). Experiments showed that HoPIMs applying \pso\ and \ga\ are competitive, while those running SSA gave the best speed-ups but computed the worst-accuracy solutions.   Finally, Silveira {\em et al.} \cite{Lucas2021} proposed a variety of HePIMs to deal with URD. In this work, we select Lucas  {\em et al.}  models \htreeDA\ and \hgbmmDA\  for our comparison with the new reconfigurable HePIMs.  

HePIMs also have been conceived to solve multiobjective optimization problems (MOPs). We believe that MOPs are exciting applications to our reconfigurable HePIMs since each island may update its BA to optimize a single objective function.  
For example, Zang \emph{et al.} \cite{Zang2011} proposed a  multi-swarm optimizer that handles each objective function of a MOP with a different slave swarm, and a master swarm covers gaps among non-dominated optima using a multiobjective \pso. Also, Xu \emph{et al.} \cite{Xu2018} proposed a model with EAs using two subpopulations to solve dynamic interval MOPs, which are MOPs that change interval parameters of their objectives or constraints over time. In addition, Gong \emph{et al.} \cite{Gong2020} proposed a model that handles a cooperative co-evolutionary MOP based on dynamic interval similarity. Gong \emph{et al.} approach split decision variables according to their interval similarity and interval parameters. Then,  decision variables are optimized cooperatively. Furthermore, Hashimoto \emph{et al.}
 \cite{Hashimoto2018} proposed a HePIM to solve multi-task problems, where each island evaluates an objective. Migrants are selected at high migration frequency, and removed randomly on each local island, replacing the worst individuals in the target islands. Since immigrants went to islands responsible for different objectives, their fitness values are the worst, assuming they have fewer chances of being suitable for the target island objective.

\section{Conclusions and future work}\label{sec:conclusion}

Reconfigurable heterogeneous PIMs were introduced. Such architectures can run and dynamically update different bio-inspired algorithms on their islands.   Two reconfigurable PIM architectures with two different archipelago-topologies were designed: a static binary tree topology, \rhtreeDA, and a dynamic complete graph topology, \rhgbmmDA. The asynchronous models ran four different BAs in their islands: \ga, \gad, \pso, and \de. 
The new reconfigurable HePIMs were tested over the unsigned reversal distance problem and computed results that outperformed the quality of associated non-reconfigurable HePIMs.   

Experiments,  evaluated statistically, made evident the potential of reconfigurable HePIMs. Such reconfigurable models preserve the power of heterogeneous PIMs to navigate efficiently in the space of feasible solutions through a healthy balance between individual and island diversity and migration policy. Also, the new reconfiguration feature gives the architecture the flexibility to evolve dynamically, improving the model's algorithmic adaptability to solve the target problem.    

The architectures \rhtreeDA\ and \rhgbmmDA\ reached quality results very competitive regarding non-reconfigurable associated architectures \htreeDA\ and \hgbmmDA, closing the gap with the homogeneous best-adapted model that uses \de. 

Future work will explore reconfigurable HePIMs over different problems and with a greater variety of BAs. In particular, we believe that the dynamic algorithmic heterogeneity promoted by the new model will be helpful to deal with multiobjective optimization problems. Indeed, the reconfiguration would promote the application of the best-adapted BA to each target problem over the islands.

%Bibliography 
\bibliographystyle{ieeetr}
\bibliography{references}
\end{document}